%% file: egpaper.tex
\documentclass[10pt,twocolumn,letterpaper]{article}

\usepackage{wacv}
\usepackage{times}
\usepackage{epsfig}
\usepackage{graphicx}
\usepackage{amsmath}
\usepackage{amssymb}
\usepackage{multirow}
\usepackage{comment}

\usepackage{bibunits}
\defaultbibliographystyle{ieee_fullname}
\defaultbibliography{egbib.bib,egbib_SUPP.bib}

\usepackage{url}

\usepackage{amsmath,amssymb,amsfonts}

\usepackage[ruled,vlined ]{algorithm2e}
\usepackage{algorithmic}

\usepackage{titling}

\def\wacvPaperID{884} 

\wacvfinalcopy 

\ifwacvfinal
\def\assignedStartPage{1} 
\fi

\ifwacvfinal
\usepackage[breaklinks=true,bookmarks=false]{hyperref}
\else
\usepackage[pagebackref=true,breaklinks=true,colorlinks,bookmarks=false]{hyperref}
\fi

\def\showpagesforarxiv{} 

\ifwacvfinal
    \ifdefined\showpagesforarxiv
    \else
    \fi
\else
\fi

\begin{document}
\title{Temporal Stochastic Softmax for 3D CNNs: \\ 
An Application in Facial Expression Recognition}


\author{
Théo Ayral$^1$, Marco Pedersoli$^1$, Simon Bacon$^2$, and  Eric Granger$^1$ \\
 $^1$ LIVIA, Dept. of Systems Engineering, École de technologie supérieure, Montreal, Canada \\
$^2$ Dept. of Health, Kinesiology \& Applied Physiology, Concordia University, Montreal, Canada\\
{\tt\small theo.ayral.1@ens.etsmtl.ca}, {\tt\small \{marco.pedersoli, eric.granger\}@etsmtl.ca}\\
{\tt\small simon.bacon@concordia.ca}
}

\makeatletter
\def\@maketitle
   {
   \newpage
   \null
   \vskip .375in
   \begin{center}
      {\Large \bf \@title \par}
      \vspace*{24pt}
      {
      \large
      \lineskip .5em
      \begin{tabular}[t]{c}
         \ifwacvfinal\@author\else Anonymous WACV submission\\
         \vspace*{1pt}\\
Paper ID \wacvPaperID \fi
      \end{tabular}
      \par
      }
      \vskip .5em
      \vspace*{12pt}
   \end{center}
   }
\makeatother

\maketitle

\ifwacvfinal
    \ifdefined\showpagesforarxiv
    \else
        \thispagestyle{empty}
    \fi
\fi

\begin{abstract}
\input{./text/abstract.tex}

\end{abstract}


\begin{bibunit}

\section{Introduction} 
\input{./text/introduction.tex}

\section{Related Work} \label{sec:related_work}
\input{./text/related_work.tex}

\section{Proposed Approach}
\input{./text/Sampling_and_Pooling.tex}
\input{./text/weighted_training.tex}

\section{Results and Discussion} \label{sec:experiments}

\input{./text/experimental_methodology.tex}
\input{./text/results.tex}

\section{Conclusion}

\input{./text/conclusion.tex}

\section*{Acknowledgements}
This work was supported by 
the Natural Sciences and Engineering Research Council of Canada (RGPIN-2018-04825), 
Calcul Québec and Compute Canada,
and 
the Canadian Institutes of Health Research (SMC-151518).

\pagebreak

{
\small

\input{egpaper.bbl}
}
\end{bibunit}


\input{egpaper_supp.tex}

\end{document}

%% file: text/abstract.tex


Training deep learning models for accurate spatiotemporal recognition of facial expressions in videos requires significant computational resources. For practical reasons, 3D Convolutional Neural Networks (3D CNNs) are usually trained with relatively short clips randomly extracted from videos. However, such uniform sampling is generally sub-optimal because equal importance is assigned to each temporal clip. In this paper, we present a strategy for efficient video-based training of 3D CNNs. It relies on softmax temporal pooling and a weighted sampling mechanism to select the most relevant training clips. The proposed softmax strategy provides several advantages – a reduced computational complexity due to efficient clip sampling, and an improved accuracy since temporal weighting focuses on more relevant clips during both training and inference. Experimental results obtained with the proposed method on several facial expression recognition benchmarks show the benefits of focusing on more informative clips in training videos. In particular, our approach improves performance and computational cost by reducing the impact of inaccurate trimming and coarse annotation of videos, and heterogeneous distribution of visual information across time.

%% file: text/introduction.tex
Deep Learning (DL) models have been successfully applied in many visual recognition tasks, including detection, classification, tracking, and segmentation, and currently achieve state-of-the-art (SOTA) performance on several image-based benchmarks \cite{DBLP:conf/cvpr/CarreiraZ17, DBLP:conf/cvpr/KarpathyTSLSF14}. Spatiotemporal recognition, in which appearance and motion features play a complementary role, remains a challenging problem in real-world applications. While many DL models are based on spatial feature extraction, specialized mechanisms are needed to manage spatiotemporal data. 

In facial expression recognition (FER), the output produced by a 2D~Convolutional Neural Network (CNN), \eg~VGG or ResNet models, in response to a sequence of frames is typically aggregated or processed by a recurrent neural network which, as seen in AVEC and EmotiW competitions~\cite{DBLP:conf/icmi/Dhall19}, can provide high-level performance~\cite{DBLP:conf/icmi/LiZZLTJLX19,DBLP:conf/icmi/LiuTLW18,DBLP:conf/icmi/LuZLTLYZ18}. 
In contrast, 3D~CNNs can process a clip as a single input, and jointly analyze appearance and motion to encode spatiotemporal relationships~\cite{de2020deep, DBLP:conf/icassp/MeloGL20, praveen2019deep}. 3D~CNNs have also been integrated as components in FER systems, not always performing well on their own with limited data~\cite{DBLP:conf/icmi/FanLLL16, DBLP:conf/icmi/LuZLTLYZ18, DBLP:conf/icmi/VielzeufPJ17}. However, recent studies have shown the relevance of 3D~CNNs for video recognition, highlighting the importance of adopting appropriate training strategies, and integrating extra training data through transfer learning~\cite{DBLP:conf/cvpr/CarreiraZ17,DBLP:conf/cvpr/HaraKS18,DBLP:conf/eccv/XieSHTM18}. Along these lines, our work is focused on efficient training strategies for 3D~CNNs.

The computational requirements of 3D~CNNs represent an important challenge in video recognition applications. Motion adds an extra dimension to model representations (\ie inputs and feature tensors are much larger), and significantly increases the computational and GPU-memory requirements for training a DL model. 
To address this issue, state-of-the-art 3D models~\cite{DBLP:conf/cvpr/CarreiraZ17,DBLP:conf/cvpr/HaraKS18} are trained with short, randomly sampled training clips, which is a stochastic approximation of temporal average pooling (see Section~\ref{sec:pooling_and_sampling}). At inference time, as memory requirements are reduced, temporal average pooling is used. 
Training with short clips has the advantage of mitigating issues related to GPU memory and video length. The efficiency of this technique in practice suggests that modeling long-range temporal dependencies is not needed in most cases to achieve accurate spatiotemporal recognition. 
However, a uniform selection of training clips from real-world videos, enforcing equal importance to all frames, raises other issues for training and inference.
Clips extracted from a video captured ``in the wild” are not all equally relevant, because of inherent characteristics of the tasks or noise in capture conditions. Assigning a global video label to short clips generates noise, and some clips can even be misleading because they do not represent the general aspect of the video. This has important implications for both training and testing phases. 
For instance, in FER applications, expressions captured in videos vary significantly depending on subjects and capture conditions (\eg illumination and pose). As a result, parts of a video may not contain any relevant information. Also, the expression intensity in FER videos varies through different states (typically onset, apex and offset~\cite{DBLP:journals/prl/KamarolJKPP17, DBLP:conf/eccv/ZhaoLLLHVY16}), and not all these states provide the same discriminative power for spatiotemporal recognition. Moreover, most of the video is typically dominated by neutral state and does not correspond to the sequence-level expression label. To avoid investing computational resources on training with uninformative clips, and to reduce the performance limitation incurred by training on incorrectly labeled clips, it is preferable to learn to sample the most relevant clips. 



\paragraph{Contribution.} In this paper, we present a new temporal stochastic softmax method to efficiently train 3D~CNNs for spatiotemporal recognition, with videos of arbitrary length and sequence-level labels. This method leverages a stochastic approximation of softmax temporal pooling for efficient sampling and learning of relevant training clips. Softmax sampling weights are estimated iteratively during training, with lower variance than the REINFORCE method~\cite{DBLP:journals/ml/Williams92}, thereby leading to better results. Although uniform clip sampling is often used for its simplicity, empirical results on several FER datasets show that the proposed temporal stochastic softmax provides a better cost-effective training approach for 3D~CNNs, achieving a higher level of accuracy, and a shorter training time.

%% file: text/related_work.tex
 
\subsection{Spatiotemporal models for FER}
 \label{sec:RW_spatiotemp}

Given the high level of performance achieved on visual recognition tasks, 2D~CNNs have been applied to video classification. These models extract a hierarchy of spatial features from each RGB frame independently~\cite{DBLP:conf/cvpr/KarpathyTSLSF14}. The resulting set of frame-level representations are then aggregated to summarize the entire video. 
Although this approach can already provide good recognition accuracy~\cite{DBLP:conf/icmi/BargalBCZ16,DBLP:conf/icmi/VielzeufKPLJ18}, it does not leverage temporal information in the data representation. Better suited methods have been proposed for video classification.
Recurrent neural networks (RNNs) are commonly used to recognize temporal patterns in the sequence of high-level frame representations produced by a 2D~CNN~\cite{DBLP:conf/cvpr/DonahueHGRVDS15,DBLP:conf/icmi/KahouMKMP15,DBLP:conf/cvpr/NgHVVMT15}.
Alternatively, two-stream convolutional networks~\cite{DBLP:conf/nips/SimonyanZ14} explicitly complement the appearance analysis, augmenting still RGB frames with precomputed optical-flow stacks describing low-level motion between frames~\cite{DBLP:conf/dagm/Sevilla-LaraLGJ18,DBLP:journals/pami/VarolLS18}. 
3D~CNNs~\cite{DBLP:conf/eccv/TaylorFLB10,DBLP:conf/iccv/TranBFTP15} unify the temporal and spatial analysis, treating videos as data volumes. 3D convolutional layers extract a hierarchy of spatiotemporal features from the RGB frame sequence. 
These powerful models are heavier to operate, because of their high number of parameters, requiring more training data and computational resources. 
Even so, 3D models are useful in FER. As an example, the small C3D~\cite{DBLP:conf/iccv/TranBFTP15} helps for the classification of relatively small video datasets, and can even be used as a deep spatiotemporal feature extractor, combined with RNNs as in~\cite{DBLP:journals/corr/abs-1804-08348}.


On action recognition, results from Carreira and Zisserman~\cite{DBLP:conf/cvpr/CarreiraZ17} and Hara \etal~\cite{DBLP:conf/cvpr/HaraKS18} show that 3D models integrating efficient transfer-learning, or simply pretrained on recent large video datasets perform better than 2D~CNNs combined with RNNs. 
Indeed, the progressive augmentation in available training data was followed by huge improvements in the performance of 3D models on action recognition tasks~\cite{DBLP:journals/corr/KayCSZHVVGBNSZ17}.
However, in the context of facial expression recognition (FER), datasets like Acted Facial Expressions in the Wild~\cite{DBLP:journals/ieeemm/DhallGLG12} (AFEW) are still smaller by an order of magnitude, limiting the performance of 3D~CNNs. Leveraging appearance and motion analysis jointly is still an issue~\cite{DBLP:conf/cvpr/DibaFSKAYG18}, although spatiotemporal methods have been studied for a long time~\cite{DBLP:conf/icml/JiXYY10}. 
In the EmotiW 2017 challenge (based on the AFEW dataset), the second place was achieved without using spatiotemporal features~\cite{DBLP:conf/fgr/KnyazevSEK18}. This direction was followed by the OL\_UC team of Vielzeuf \etal~\cite{DBLP:conf/icmi/VielzeufKPLJ18} who made a point of not using any motion features for simplicity and efficiency.






However, experiments in the study of Valstar and Pantic~\cite{DBLP:journals/tsmc/ValstarP12} suggest that temporal analysis should play a role for FER, not only on the high-level description of expression dynamics throughout the video but also in the detection of local motion features. 
In the more general context of action recognition,  Sevilla-Lara \etal~\cite{DBLP:journals/corr/abs-1907-08340} and Huang \etal~\cite{DBLP:conf/cvpr/HuangRMTPFN18} evaluated the importance of motion analysis, and the ability of state-of-the-art models to capture it. 
They also hypothesized that long-term patterns are not necessary for action recognition, but a wide receptive field helps to capture the most relevant frames~\cite{DBLP:conf/aaai/LiuLGTM18}. 
Our work is in line with these considerations, as the proposed stochastic softmax consists of an efficient weighting of short video clips for 3D~CNNs.

\subsection{Efficient weighting of clips}


Works related to ours are ones that study techniques to improve training with better sample selection, or to improve inference using a weighted aggregation method. Temporal aggregation of features for 3D~CNNs is usually performed with average pooling~\cite{DBLP:conf/cvpr/CarreiraZ17,DBLP:conf/cvpr/HaraKS18,DBLP:conf/cvpr/DibaFSKAYG18,DBLP:journals/pami/VarolLS18}
(more details are provided in Section~\ref{sec:pooling_and_sampling}).
The theoretical analysis of Boureau \etal~\cite{DBLP:conf/icml/BoureauPL10}, as well as experimental results~\cite{DBLP:journals/taslp/McFeeSB18}, shows that max and average pooling have different ranges of expertise, depending on the input size and the sparsity of features. To help find the most adapted type of pooling, softmax is proposed as a parameterizable generalization of these two pooling techniques.

The pooling strategy used for temporal feature aggregation also has a great influence on training, by deciding which part of the input will generate gradients. For 3D~CNNs, as training is performed on short clips (usually 8 to 64 frames), the receptive field of the aggregation mechanism is limited, and the pooled features have a different distribution than at test time, with longer videos. Consequently, other weighting methods have to be developed. 
Studies on importance sampling~\cite{DBLP:conf/icml/KatharopoulosF18,DBLP:conf/cvpr/WangSLRWPCW14} have made clear that all samples do not have the same relevance to the training process. 
However, for most video classification datasets (\eg AFEW~\cite{DBLP:journals/ieeemm/DhallGLG12} and  Kinetics~\cite{DBLP:journals/corr/KayCSZHVVGBNSZ17}), labels are not available for each of the frames or clips, but only at the video level. 
Related issues are discussed by Zhu \etal~\cite{DBLP:conf/cvpr/ZhuHSCQ16}, as they consider short-clip training as a weakly supervised learning problem within the action recognition task with video-level labels. 
In this multiple instance learning (MIL) framework, a bag of several clips is fed to the model and temporal max pooling is used to learn only from the best scored clip. This is a way of providing better training data by selecting the most informative temporal windows \textit{a posteriori}. This approach still requires the use of several clips per sample at each epoch, which is problematic for 3D models. To be able to select clips before evaluating them with the entire model, this method is complemented with a motion metric computed offline. In the context of action recognition, the hypothesis is that relevant clips are the ones with more motion. This does not hold for FER, so our method is purely based on classification scores, and we compute them online with a single training-clip per sample for each epoch, to save computation with the 3D~CNN. 

Also to cope with capture, trimming and labelling noise, the weighted C3D~\cite{DBLP:conf/icmi/VielzeufPJ17} integrates a softmax layer to give more importance to relevant clips during training. All windows are evaluated in the early epochs of training and their scores are used to weight the training loss, reducing the effect of uninformative or wrongly labeled clips on the model parameter updates.
%
The weighting strategy is amplified throughout training, from average to max. 
This principle of weighted training constitutes a basis of our work, with the idea of identifying relevant training clips for 3D~CNNs. Yet we replace the loss weighting by a stochastic sampling  mechanism to avoid computing gradients that would be inhibited by the softmax, and we rethink the distribution estimation method to remove the computational overhead of evaluation. This allows us to train a model with more parameters than the small C3D.

\pagebreak

Another interesting mechanism is the SCSampler~\cite{DBLP:conf/iccv/KorbarTT19}, trained to select the best clips to feed to the classifier at test time. This external and light-weight sampler is trained to predict the relative saliency of clips. 
By running the classifier on a few sampled clips instead of the entire video, the computational cost at inference time is reduced and classification accuracy is improved. 
This study shows experimentally that the classification score is a good metric of informativeness of a clip. 
Our approach is related to their Oracle Sampler, 
which ranks clips according to their scores for the target class.
We adapt this mechanism for training-clip selection, and propose an approach to avoid dense application of the classifier through iterative sampling (Section~\ref{sec:implementation}).

%% file: text/Sampling_and_Pooling.tex



The main objective of this work is to perform a pooling operation that can select the most important frames of a video sequence. As presented in Section~\ref{sec:RW_spatiotemp}, softmax pooling seems to be the right candidate because it assigns a specific weight to each short clip of the video. However, standard softmax pooling requires an evaluation of all short clips of a video sequence. This is unfeasible for large 3D models as they require a large amount of memory and computation. 
In this section we show how to approximate softmax pooling during training such that it requires the evaluation of only one short-clip per video at each training iteration. This makes the training much lighter, while optimizing the same objective function in expectation.

\begin{figure}
\centering
\includegraphics[width=1\linewidth]{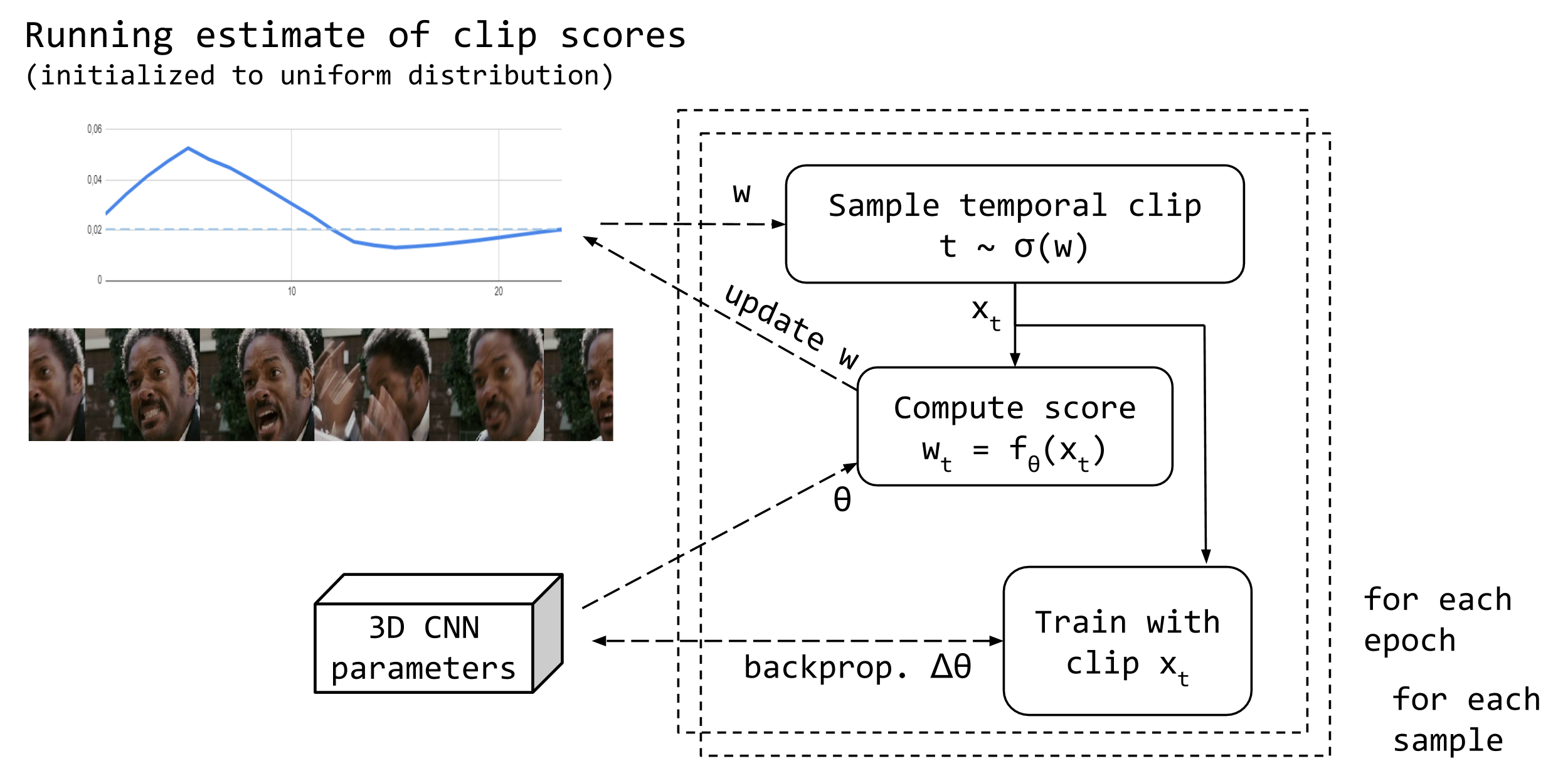}

\caption{Illustration of weighted clip sampling for temporal stochastic softmax training. Within each video, training clips are sampled based on the softmax of their classification-score distribution. Running estimates of clip scores are updated at every iteration with the classification score of the selected clip.}
\label{fig:pipeline}
\end{figure}

An overview of the stochastic softmax training process is shown in Figure~\ref{fig:pipeline}.
In Section~\ref{sec:pooling_and_sampling} we present the motivations and challenges of integrating softmax pooling into the usual short-clip training framework. Then, Section~\ref{sec:distribution_estimation} discusses solutions for estimating temporal probability distributions iteratively from short clips. Finally, Section~\ref{sec:implementation} provides the details of our stochastic softmax pooling, combining both training-clip sampling and softmax pooling.

\subsection{Softmax pooling with short-clip sampling}
\label{sec:pooling_and_sampling}



Our objective is to minimize the loss $\mathcal L$ of a parameterized classifier $f$ on the training dataset. For each video $x$ with label $y$, we minimize $\mathcal{L}(f(x),y)$. For video classification, a typical example of such loss is cross-entropy. If using temporal average pooling, the loss can be written as $\mathcal{L}(\frac{1}{T}\sum^T_{t=1} f_t(x),y)$ in which $f_t(x)$ represents the learned features associated to 
temporal position $t$ in a video of duration $T$.
Assuming that the temporal receptive field of the network is limited or that the features extracted for time $t$ only depend on a small temporal neighbourhood (a clip), we can compute the loss as $\mathcal{L}(\frac{1}{T}\sum^T_{t=1} f(x_t),y)$ where $x_t$ is a clip associated to time $t$. This assumption is implicitly or explicitly used in most of the recent work on 3D~CNNs for video classification~\cite{DBLP:conf/cvpr/CarreiraZ17,DBLP:conf/cvpr/HaraKS18,DBLP:conf/cvpr/DibaFSKAYG18,DBLP:journals/pami/VarolLS18} because it enables the use of an 
approximation of the loss: 
\begin{equation}
    \mathcal{L}( \frac{1}{T} \sum^T_{j=1} f(x_j),y ) 
    \approx 
    \mathcal{L}(f(x_t),y), \hspace{.3cm} t \sim \mathcal{U}(1,T).
    \label{equ:approx}
\end{equation}

The loss is computed by sampling different clips throughout training. For each iteration, the loss of a video is approximated using a single clip that is uniformly sampled from each video. This produces significant reduction in computational complexity of each training step and in GPU memory requirements necessary to make the training of 3D~CNNs possible. 



Here we show that this sampling technique with cross-entropy loss is an upper-bound of the real loss. 
Indeed, cross-entropy loss is convex, and based on Jensen's inequality, the averaged loss computed on all clips $\frac{1}{T}\sum_{t=1}^T\mathcal{L}(f(x_t),y)$ is an upper-bound of the cross-entropy of the averaged-pooled features
$\mathcal{L}(\frac{1}{T}\sum^T_{t=1} f(x_t),y)$:  
%
\begin{equation}
\hspace{-0.2cm}\frac{1}{T}\sum^T_{t=1}\mathcal{L}(f(x_t),y)
\ge
\mathcal{L}(\frac{1}{T}\sum^T_{t=1}f(x_t),y)
=
\mathcal{L}(f(x),y).\hspace{-0.1cm}
\label{equ:jensen_uniform}
\end{equation}
%
The average of clip-level losses is an empirical estimation of the expected loss $\mathop{\mathbb{E}}[\mathcal{L}(f(x_t),y)]$. Instead of computing the entire sum, the expected loss is approximated with a single sample $t$ uniformly sampled as in Eq.~(\ref{equ:approx}). Thus, in expectation, the same loss is optimized during training. At test time, 
memory is less problematic as gradients are not computed, so inference is computed as the average of all video clips, thus a temporal average pooling.

Although this form of training with uniformly sampled clips can be effective, it restricts the temporal pooling to the average pooling strategy. Our work addresses this issue by proposing the more general softmax strategy in the training-clip sampling framework. 
Temporal average pooling assumes that each sub-region (clip for videos) of the pooled features contains information that is important for the task. 
 It is expected to perform well when videos are short and entail exactly the category that we want to classify~\cite{DBLP:conf/icml/BoureauPL10,DBLP:journals/taslp/McFeeSB18}. On the other hand, when a video is longer, complex and with possible noise, max pooling is expected to work better. In general, the optimal level of importance for the different parts of a video is unknown, as it depends on the task and the actual data. 
In this paper, we propose to use weighted pooling in which a weight $p_t$ is associated with each temporal position $t$ of the video.
This resembles an attention mechanism~\cite{DBLP:conf/icml/XuBKCCSZB15,DBLP:conf/iciar/AminbeidokhtiPC19,DBLP:conf/cvpr/GirdharCDZ19,DBLP:journals/pr/LiLZW20,DBLP:conf/icip/MengPW019}, but instead of estimating attention weights with a learned layer, $p_t$ is computed as a temporal softmax of the score associated to a given clip.  
The relative importance attributed to each clip depends on its classification score:
\begin{equation}
\hspace{-0.15cm}    f(x) = \sum^T_{t=1} \frac{\exp(\gamma f(x_t))}{\sum_{j=1}^T  \exp(\gamma f(x_j))} f(x_t) = \sum^T_{t=1} p_t f(x_t). \hspace{-0.05cm}
    \label{equ:softmax}
\end{equation}
The softmax operator is parameterized by $\gamma$, the inverse temperature. When $\gamma = 0$, the weight vector $p = (p_1, p_2, ... p_T)$ is equal to the center of the unit simplex, with $p_t = \frac{1}{T} \:\forall t$, so the operator is equivalent to average pooling. In contrast, when $\gamma \xrightarrow{} +\infty$ all the weight will be assigned to the highest scoring $t$, as in max pooling. Equation~(\ref{equ:softmax}) is therefore a generalization of max and average pooling, parameterized by a factor $\gamma$. 
As with uniform clip sampling, we 
provide an upper-bound on the training loss obtained with softmax pooling: 
\begin{equation}
\hspace{-0.15cm}p_t\sum^T_{t=1}\mathcal{L}(f(x_t),y)
\ge
\mathcal{L}(p_t\sum^T_{t=1}f(x_t),y)
=
\mathcal{L}(f(x),y).\hspace{-0.05cm}
\label{equ:jennsen_softmax}
\end{equation}
The weighting factor $p_t$ of softmax pooling, from Eq.~(\ref{equ:softmax}), becomes a sampling probability distribution in softmax short-clip training. Indeed, for each video sample $x$, instead of weighting losses obtained from clips sampled uniformly, we directly weight the sampling distributions of clips. In this way we select clips that are more important for training than with uniform sampling.

%% file: text/weighted_training.tex
\subsection{Estimation of the sampling distributions}
\label{sec:distribution_estimation}

In Eq.~(\ref{equ:softmax}), we see that in order to compute $p_t$, $f(x_t)$ must be evaluated on all clips $t$ of a video, which is computationally expensive and consumes considerable memory. We seek to avoid this issue with a stochastic sampling strategy.
Thus, instead of computing $p$ as in Eq.~(\ref{equ:softmax}), we introduce a new variable $q = (q_1,q_1,...,q_T)$ that estimates $p$ for each video $x$. 
During training, for each video, $q$ is defined 
as minimizing the loss. 
A straightforward way to estimate 
$q$ is by using REINFORCE~\cite{DBLP:journals/ml/Williams92}. We consider the loss as an expectation over time, sampled with $q$.
Thus, its gradient will be:

\begin{equation}
\label{equ:grad_reinforce}
    \nabla_q \mathop{\mathbb{E}}_{t \sim q}[\mathcal{L}(f(x_t),y)] 
    = \mathop{\mathbb{E}}_{t \sim q}[\mathcal{L}(f(x_t),y) \nabla_q log(q_t)]. 
\end{equation}

\clearpage

%

Unfortunately, the gradients estimated with REINFORCE have high variance and, even when using a baseline of the expected cumulative reward, the updates of $q_t$ are too noisy. The potential benefits of sampling are therefore lost by a poor estimation of~$q$. Results and limitations of the estimation of sampling distributions with REINFORCE are discussed in Section~\ref{sec:ablation_study}. As the distribution parameters are essentially pushed to favour the selection of high scoring (low loss) training clips, it is possible to ``shortcut" the REINFORCE optimization by 
building the sampling distributions directly from the clip scores.
Thus, 
we propose to estimate $q$ in a close form, without the use of gradients. Since $p_t$ is calculated as softmax of $\gamma f(x_t)$, it is possible to store the values of 
$w_{x,t} = f(x_t)$ directly, and apply softmax when an estimation of $q$ is needed to select the training clip. 
The probability distribution $q$ is therefore computed from running estimates of scores evaluated at different iterations. This approach is quite simple, does not rely on a noisy estimation of the gradients, and works well in practice. 
During training, softmax sampling maximizes the classification score for the correct label (providing relevant clips to the model), while keeping diversity in clip sampling to avoid overfitting~\cite{DBLP:journals/corr/abs-1704-00805}.


\subsection{Implementation of stochastic softmax} 
\label{sec:implementation}

Videos are classified by convolving the 3D~CNN in the temporal dimension, over all possible overlapping windows, and implementing a temporal pooling mechanism on the resulting clip-level scores~\cite{DBLP:conf/cvpr/CarreiraZ17}. We combine this evaluation scheme with single-clip extraction during training, as only one clip is used from each video at every epoch. 
For more details on the implementation, refer to the Supplementary Material.

\paragraph{Clip Sampling.}

The sampler $S$, extracts a clip of $F$ contiguous frames at temporal position $t$ from a video $x$ of arbitrary length $L$. At every epoch of training, we construct batches of training clips. One clip is sampled from each training video. Let $w_x$ be the temporal sequence of $N=L-F+1$ classification-score estimates corresponding to the temporal responses of the classifier convolved over $x$. Then $w_{x,t}$ is the estimated classification score for training clip $x_t$. This score will be the base of our clip weighting. Temporal softmax sampling follows the formula:
\begin{equation}
\label{eq:weighted_sampling}
p(S(x)=x_t) = \frac{\exp(\gamma w_{x,t})}{
\sum_{n=1}^{N} 
\exp(\gamma w_{x,n})} \; , \hspace{.3cm} \forall x_t \subset x .
\end{equation}

The principle of stochastic softmax training is summarized in Eq.~(\ref{eq:weighted_sampling}). At test time, the relative importance attributed to each clip through temporal pooling is defined as the weighting factor computed from the softmax function in Eq.~(\ref{equ:softmax}). During training, the temporal weighting is implemented as a sampling probability which translates into a frequency of occurrence in the training iterations. Instead of classifying entire videos and applying temporal weights to the loss afterward, computation is exclusively focused on the selected clip.

\paragraph{Distribution updates.} 

The proposed approach stores running estimates $w$ of the score distribution of each training video, and updates them 
at every epoch, as video clips are sampled. Through iterative clip sampling, we obtain information on the temporal classification score distribution of each video and use it to build its temporal clip selection probability distribution. An overview of the training process is presented in Algorithm~\ref{algo:weighted_training}.

\begin{algorithm}
\SetAlgoLined
 Initialize $w$ with uniform distributions\;
 \ForEach{epoch}{
 \ForEach{video $x$ in training set}{
    compute the clip sampling distribution from~$w_x$ with Eq.~(\ref{eq:weighted_sampling}) \;
    sample clip $x_t$, with $t=S(w_x)$ \;
    compute the score for the correct class $y$ : 
    $w_{x,t} = f(x_t)[y]$ \;
    update $w_x$ around sampled location $t$ \;
	train with clip $x_t$ by back-propagation \;
 }
 }
 \caption{Stochastic softmax training}
 \label{algo:weighted_training}
\end{algorithm}

\paragraph{Training phases.}
Efficient training-clip sampling is highly dependent on the accuracy of the temporal distributions. 
Therefore, we implement several mechanisms to bootstrap the distributions, to make them representative of the informativeness of clips as early as possible during training, without introducing heavy computational overhead. We decompose the training process into three simple steps relative to the sampling mechanism: first, warm-up with uniform sampling and no distribution updates, then, exploration with deterministic sampling and initialization of distributions, and finally exploitation with softmax weighted sampling and distribution updates as described above.

%% file: text/experimental_methodology.tex
In this section, we perform an ablation study of the method on emotion recognition with AFEW and validate on two datasets of pain detection to evaluate the generalization power in the scope of facial expression recognition. This section only contains the most important results and illustrations to support the paper. For additional experimental results, see the Supplementary Material.

\subsection{Experimental methodology}

\paragraph{AFEW.}

The Acted Facial Expressions in the Wild dataset of emotion recognition~\cite{DBLP:journals/ieeemm/DhallGLG12} was used to evaluate the proposed and reference training methods. The task is to classify video samples by assigning each of them a single emotion label from the six universal emotions (Anger, Disgust, Fear, Happiness, Sad \& Surprise) and Neutral. The performance metric is the classification accuracy (video-level rank-1 accuracy). We do not consider the audio information in our study. The Training set contains 773 samples from 67 movies, with 228 actors. The Validation set contains 383 samples from 33 movies, with 134 actors. The dataset has been constructed in a subject independent manner. AFEW video duration ranges from 0.6s to 5.4s (between 16 and 128 frames), with an average of 2.5s. When using the SeetaFace Engine to crop and align faces~\cite{DBLP:journals/ijon/WuKHSC17}, the average bounding box of source crop has size 265$\times$265 (from the original 720$\times$576 image).
Models are evaluated on the Validation set of AFEW (the Test set being reserved for the EmotiW competition). Model training and hyper-parameter validation were performed on random splits of the Training set. Created from movie samples, AFEW provides close to real-world data, with a wide range of challenges due to the variation in head poses and movements, illumination and backgrounds. Additional noise comes from camera motion, sometimes causing occlusion. Some of these issues are well addressed by the face alignment process. Others can be managed with the proposed temporal weighting mechanism. The dataset was created with a semi-automatic extraction process, producing an imprecise trimming of movie samples. Videos are not always centered on the relevant emotional frames, different scenes can be present in a video, and multiple subjects can be present in the same frame.

\paragraph{UNBC-McMaster.}

The UNBC-McMaster Shoulder Pain database 
\cite{DBLP:conf/fgr/LuceyCPSM11}
contains 200 image sequences capturing the spontaneous pain expressions of 25 subjects. Sequences vary in length from 48 to more than 500 frames.
We follow Wu \etal~\cite{DBLP:conf/fgr/WuWJ15} for the evaluation task. The dataset is used in a binary classification setup based on the Observed Pain Intensity (OPI) expert annotations. The 92 sequences with OPI = 0 constitute the negative samples (No Pain), while the Pain class is composed of the 57 sequences with OPI $\geq$ 3. Because of the class imbalance, the evaluation metric used for this task is the classification accuracy at Equal Error Rate on the Receiver Operating Characteristic curve (ROC-EER). We perform leave-one-subject-out cross-validation for the 25 subjects.

\paragraph{BioVid.}

The BioVid Heat Pain Database~\cite{DBLP:conf/bmvc/WernerAN0GT13}, Part A, is a relatively large heat-pain detection dataset, with 20 frontal video recordings per stimulus level, for each of 87 subjects. 
Participants received four levels of painful stimuli (PA1 to PA4), adapted to their subject-specific sensitivity. Videos with no stimulus are also present to constitue the BL1 class (No Pain). Biomedical signals are also available but not used in our study. As opposed to UNBC-McMaster, BioVid provides objective labels based on the temperature of the heat-pain inducing device (with subject-specific levels). The task is thus much more difficult, as the objective is not to classify the directly observable expression but its source. All subjects do not react to pain with the same intensity, even though the stimuli are calibrated for each participants. The creators of the database studied this phenomenon and identified participants that did not react visibly to the pain-inducing stimulus~\cite{DBLP:conf/acii/WernerA017}. The BioVid videos are even more controlled than UNBC-McMaster and contain less head-pose variations and occlusion~\cite{10.1371/journal.pone.0192767}. However, the number of irrelevant frames remains an issue, because videos are not trimmed to capture the specific expression but span a fix time window of 5.5 seconds based on the timing of the stimulus. Thus, results on BioVid should specifically evaluate the ability of training-clip selection to favour expressive windows over relatively neutral states usually present at the beginning and end of every video. On this dataset, we use two different evaluation protocols from the literature.
The first task is binary classification of Neutral (BL1) and highest level of pain (PA4), with 1740 videos per class. We perform 8-fold subject-independent cross-validation and report classification accuracy. The second setup is similar but the Pain class is extended to PA3 and PA4 videos, as proposed by Yang \etal~\cite{DBLP:conf/ipta/YangTLBPFH16}. To work with the class imbalance (1740 No Pain and 3480 Pain videos), Area Under the Receiver Operating Characteristic Curve (ROC-AUC) is used as metric.

For all datasets, our stochastic softmax pooling only requires sequences of RGB frames and the corresponding sequence-level class labels. Facial expression recognition requires a pre-processing step consisting of detection and alignment of faces in each video frame. We employed the SeetaFace Engine from Wu \etal~\cite{DBLP:journals/ijon/WuKHSC17}.

\paragraph{Architectures.}
We evaluate our method with an inflated~\cite{DBLP:conf/cvpr/CarreiraZ17} VGG-16 model~\cite{DBLP:journals/corr/SimonyanZ14a}. Prior to inflation, the model was pretrained with VGG-Face and FER2013 image datasets~\cite{DBLP:conf/bmvc/ParkhiVZ15,DBLP:conf/iconip/GoodfellowECCMHCTTLZRFLWASMPIPGBXRXZB13}.

In order to facilitate comparison with the literature on AFEW, we also experiment with a C3D model~\cite{DBLP:conf/iccv/TranBFTP15}, pretrained with Sports-1M~\cite{DBLP:conf/cvpr/KarpathyTSLSF14}.
We use data augmentation consistently across frames\footnote{\url{https://github.com/hassony2/torch_videovision}} for a given video, with horizontal flip, random rotation, crop and color jitter. Models are trained on a single Tesla V100 GPU, with standard SGD with momentum 0.9. We used early stopping on the validation loss. Training clips of 16 frames are selected with our stochastic softmax sampling, and temporal softmax pooling is applied to aggregate clip-level scores during inference. For the baseline, clips are selected uniformly and temporal average pooling is applied.

%% file: text/results.tex


\subsection{Ablation Study}
\label{sec:ablation_study}

\begin{figure} 
\centerline{\includegraphics[width=1\linewidth]{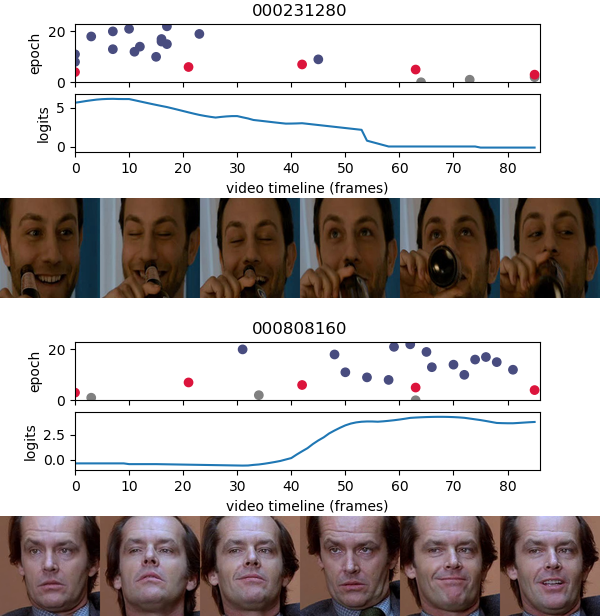}}
 
\caption{Visualization of sampling distributions (logits) and resulting clip selection during softmax training with $\gamma = 1$, for AFEW videos (Happy). The sampling maps indicate the temporal position of the selected clip at each epoch for a given video example. Colors indicate the training phase: uniform warm-up, deterministic exploration, and weighted sampling. Stochastic softmax avoids selection of occluded and neutral frames.}
\label{fig:afew_happy}
\end{figure}

\paragraph{Illustrative examples.}
Figure~\ref{fig:afew_happy} displays the sampling map and distribution of temporal stochastic softmax sampling, with two training videos from the AFEW dataset. The proposed method is shown to quickly identify the emotion intensity distribution in the video, thanks to the exploration steps at the beginning of training, and focuses on training clips with the best scores. As observed on the sample images, the video-level label ``Happy" does not correspond to the beginning and end of the videos, because of occlusion and neutral state respectively. In both cases, the model avoids these uninformative frames.

\begin{table} 
  \begin{center} 

    \begin{tabular}{|l||c|c|c|c|} 
    
      \hline 
      \centering \textbf{Inverse } 
      & \multicolumn{2}{c|}{\textbf{REINFORCE}}  
      & \multicolumn{2}{c|}{\textbf{Ours Softmax}}  \\
      
    \textbf{Temp. } 
      & \textbf{Acc.(\%)} 
      & \textbf{Ep.} 
      & \textbf{Acc.(\%)} 
      & \textbf{Ep.} \\
    \hline
    $\gamma =$ 0 & 45.66 $\pm .21$ & 24.6 
    & 45.66 $\pm .21$ & 24.6  \\
    $\gamma =$ 0.5 & 46.09 $\pm .41$ & 23.8 
    & 46.07 $\pm .27$ & 23.6 \\
    $\gamma =$ 1 & 46.80 $\pm .63$ & 22.5 
    & \textbf{47.35} $\pm .27$ & 20.3  \\
    $\gamma =$ 10 & 44.52 $\pm .18$ & 17.5 
    & 46.65 $\pm .40$ & 17.2  \\

       \hline
    \end{tabular}
  \end{center}
 
      \caption{Average performance and duration of REINFORCE and stochastic softmax training on AFEW. Both methods correspond to uniform sampling when $\gamma = 0$.}
    \label{Tab:reinf_vs_uniform}
\end{table}

\paragraph{Stochastic softmax sampling strategies.}
Table \ref{Tab:reinf_vs_uniform} compares results for stochastic softmax training and REINFORCE sampling. Rank-1 accuracy and the number of epochs needed to converge are presented when varying the softmax temperature. Inverse temperature $\gamma = 0$ corresponds to uniform sampling, as generally used in previous approaches. Higher inverse temperatures tend to approximate max pooling. As expected the best results are found in between average and max pooling. This shows that using a softmax pooling is important to achieve optimal performance on this task. A value of $\gamma = 1$ provides the best performance for REINFORCE as well as for our proposed method. However, the proposed softmax training manages to obtain better performance than REINFORCE because it has a lower variance in the estimation of $p$. Also the number of epochs needed to converge for the best temperature is reduced by around 20\% compared to uniform sampling. Focusing on the most important parts of the video not only improves the accuracy, it also allows the model to directly focus on important clips, thereby saving training time.






\subsection{Results}
\label{sec:results}
Table~\ref{tab:Afew_Literature_results} presents our results in comparison with other 3D-CNN approaches on AFEW. Our inflated VGG-16 achieves very good performance compared to other 3D~CNNs. 
Temporal stochastic softmax is able to improve accuracy further. C3D Weighted~\cite{DBLP:conf/icmi/VielzeufPJ17} performs a temporal weighting of the training-clip losses, thus a simplified approximation of our method. We also validate experimentally the effect of stochastic softmax training with a C3D, as this architecture has been extensively studied in the literature. Results show that the training method can be adapted to different architectures of 3D~CNNs.

\begin{table} 
  \begin{center} 
    \begin{tabular}{|l|l|r|} 
      \hline
      \textbf{Method} & \textbf{Model} & \textbf{Acc. (\%)} \\

    \hline  \hline

        Lu \etal, 2018 
       ~\cite{DBLP:conf/icmi/LuZLTLYZ18}
        & 3D VGG-16  & 39.36 \\
        
        \hline
        
           Fan \etal, 2016~\cite{DBLP:conf/icmi/FanLLL16} & C3D & 39.69 \\  
   

        
        \hline
 
        Vielzeuf \etal, 
        & C3D-LSTM 
        & 43.2\phantom{0} \\  
        2017~\cite{DBLP:conf/icmi/VielzeufPJ17}  
        & C3D Weighted & 42.1\phantom{0}  \\

        \hline
 
        C3D baseline 
        & C3D (uniform) &  39.95 \\ 
        C3D with Softmax 
        & C3D  ($\gamma = 1$)& 42.78 \\ 
 
        VGG baseline 
        & 3D VGG-16 (unif.)& 45.66 \\
        VGG with Softmax 
        & 3D VGG-16 ($\gamma = 1$) & 47.35 \\

        \hline

    \end{tabular}

  \end{center}
      \caption{Accuracy of a 3D~CNN with stochastic softmax  compared to our baseline (same architecture but with uniform training-clip sampling and average pooling) and relevant literature on AFEW.
    }
    \label{tab:Afew_Literature_results}
\end{table}

\begin{table} 
  \begin{center} 
    \begin{tabular} {|l|l|c|} 
      \hline 
      
      \multirow{2}{*}{\textbf{Method}}
      & \multirow{2}{*}{\textbf{Model}}
      & \textbf{EER} \\
        && \textbf{Acc. (\%)}    \\
    \hline  \hline



Wu \etal, 2015~\cite{DBLP:conf/fgr/WuWJ15} & MIL-HMM & 85.2\phantom{0} \\
\hline 

Werner \etal, 2017~\cite{DBLP:journals/taffco/WernerALWGT17}
& SPTS+CAPP & 91.7* \\
        \hline

    
Sikka and Sharma, 
&  LOMo with
& \multirow{2}{*}{87.0\phantom{0}}  \\

2018~\cite{DBLP:journals/pami/SikkaS18}&  SIFT, LBP&\\

\hline
Kumawat \etal, 
&
\multirow{2}{*}{LBVCNN} & \multirow{2}{*}{86.55}  \\

 2019~\cite{DBLP:conf/cvpr/KumawatVR19}&&  \\


        \hline

        Our baseline 
        & 3D VGG (unif.) & 86.58  \\
        Stochastic Softmax 
        & \multicolumn{1}{|r|}{" \quad ($\gamma = 2$)} & 87.21   \\

        \hline

    \multicolumn{3}{r}{\footnotesize * Accuracy is not computed at ROC-EER.}\\

    \end{tabular}

  \end{center}
      \caption{Accuracy of a 3D~CNN on UNBC-McMaster, with and without stochastic softmax, compared to related SOTA methods.
    }
    \label{tab:UNBC_results}
\end{table}

\begin{table}
  \begin{center} 
    \begin{tabular}{|l|l|c|}
      \hline 
      \textbf{Method} & \textbf{Model} & \textbf{Acc. (\%)}  \\

    \hline  \hline

Werner \etal,

& \multirow{2}{*}{Standardized FADs }& \multirow{2}{*}{72.4*} \\
2017~\cite{DBLP:journals/taffco/WernerALWGT17} & & \\

        \hline



Othman \etal, 
& RFc with FADs &  65.8\phantom{0}

\\

2019~\cite{DBLP:conf/imspa/OthmanWSA019} & Reduced MbNetV2  & 65.5\phantom{0}

\\

\hline 
Thiam \etal, 
& \multirow{2}{*}{Two-stream VGG } &\multirow{2}{*}{69.25} \\

2020~\cite{DBLP:journals/sensors/ThiamKS20} & & \\

        \hline

        Our baseline 
        & 3D VGG (uniform) & 68.12 \\
        Stochastic Softmax 
        & 3D VGG ($\gamma = 2$) & 69.60 \\

        \hline

    \multicolumn{3}{r}{\footnotesize * Using additional information with depth sensor 3D maps.}\\

    \end{tabular}             

  \end{center}

      \caption{Binary Classification accuracy of a 3D~CNN on BioVid (BL1 vs. PA4), with and without stochastic softmax, compared to related SOTA methods.}
      
    
    \label{tab:Bio1_results}
\end{table}

\begin{table}
  \begin{center} 
    \begin{tabular} {|l|l|c|c|} 
      \hline 
      \textbf{Method} & \textbf{Model} & \textbf{AUC (\%)} \\

    \hline  \hline

Tavakolian and & 3D ResNet &  82.54 \\
 Hadid, 2019~\cite{DBLP:journals/ijcv/TavakolianH19}  & S3D-G &  83.26 \\
  & SCN &  86.02 \\

        \hline

        Our baseline 
        & 3D VGG (uniform) & 82.67\\
        Stochastic Softmax 
        & 3D VGG ($\gamma = 2$) & 84.39 \\

        \hline

    \end{tabular}

  \end{center}

      \caption{ROC-AUC results on BioVid (BL1 vs. PA3-4), with and without stochastic softmax, compared to related SOTA methods.
    }
    \label{tab:Bio2_results}
\end{table}

\pagebreak

Tables~\ref{tab:UNBC_results}, \ref{tab:Bio1_results} and \ref{tab:Bio2_results} report the performance of our sampling method on three pain video classification tasks. An aggressive sampling temperature could be expected to generate overfitting on UNBC-McMaster, by reducing the variation in training data which is already very limited. Actually, we found that better results were obtained with a higher temperature, $\gamma = 2$ (Table~\ref{tab:UNBC_results}). The classification scores of the model on UNBC-McMaster are generally lower than for AFEW. The temperature parameter allows us to adapt the weighting strategy. Note that with short-clip training, the model cannot adapt its scores to learn a video-level temperature implicitly. 
Table~\ref{tab:Bio1_results} provides results on the larger BioVid dataset.
Temporal max-pooling during inference might not be beneficial, because it is necessary to consider the entirety of the video to classify it. For example, the model could recognize a neutral expression at the beginning of a Pain video, with very high confidence. We considered decoupling the temperatures of the sampling and pooling but do not extensively study this possibility here. We provide preliminary results in Supplementary Material. The good performance improvement obtained with our temporal softmax on BioVid (Table~\ref{tab:Bio1_results}) suggests that the method is also relevant to very controlled recordings, with no occlusion and very limited head movement. The benefits of clip sampling does not only consist of limiting the impact of noisy labelling or capture conditions. Temporal weighting addresses a challenge inherent to the task, as facial expressions are events localized in time, and typically preceded and followed by neutral states. This idea is confirmed as the best performance gain of temporal softmax is observed on the second task of BioVid (Table~\ref{tab:Bio2_results}). In this scenario, additional samples are gathered in the Pain class (PA3 + PA4). These are videos where temporal weighting is really efficient, as opposed to No Pain videos which do not contain apex or particularly relevant segments. This calls for future work studying the relevance of using different softmax temperatures adapted to each class, as was developed by McFee \etal~\cite{DBLP:journals/taslp/McFeeSB18} with temporal pooling for audio data.

%% file: text/conclusion.tex
We presented a softmax-based training and inference method for 3D~CNNs adaptable to the task at hand in terms of computation, regularization and feature aggregation strategy, with no additional trained layer. 
Our method is designed to enable the use of softmax temporal pooling within the framework of short-clip training, which is the standard way of training 3D models because of computational and GPU-memory limitations.
We demonstrated the benefits of softmax temperatures for video classification by considering videos as bags of unequally relevant clips. At test time, a temporal softmax pooling mechanism is able to weight and aggregate information from different clips, with a strategy adapted to the input distribution. 
Stochastic softmax sampling improves learning by balancing informative and difficult training clips, allowing for faster convergence and limiting the impact of irrelevant clips in the context of weak, sequence-level annotations. 
These mechanisms provide an improvement in accuracy on all evaluated datasets. 
Experiments suggested several directions for future work.

%% file: egpaper_supp.tex
\clearpage

\title{Temporal Stochastic Softmax for 3D CNNs\phantom{:}\\ 
– Supplementary Material –}
\author{
Théo Ayral$^1$, Marco Pedersoli$^1$, Simon Bacon$^2$, and  Eric Granger$^1$ \\
$^1$ LIVIA, Dept. of Systems Engineering, École de technologie supérieure, Montreal, Canada \\
$^2$ Dept. of Health, Kinesiology \& Applied Physiology, Concordia University, Montreal, Canada\\
{\tt\small theo.ayral.1@ens.etsmtl.ca}, {\tt\small \{marco.pedersoli, eric.granger\}@etsmtl.ca}\\
{\tt\small simon.bacon@concordia.ca}
}

\makeatletter
\def\@maketitle
   {
   \newpage
   \null
   \vskip .375in
   \begin{center}
      {\Large \bf \@title \par}
      \vspace*{24pt}
      {
      \large
      \lineskip .5em
      \begin{tabular}[t]{c}
         \ifwacvfinal\@author\else Anonymous WACV submission\\
         \vspace*{1pt}\\
Paper ID \wacvPaperID \fi
      \end{tabular}
      \par
      }
      \vskip .5em
      \vspace*{12pt}
   \end{center}
   }
\makeatother
 
\maketitle

\begin{bibunit}

\appendix

\setcounter{enumiv}{9}

\setcounter{equation}{0}
\renewcommand{\theequation}{S.\arabic{equation}}
\setcounter{figure}{0}
\renewcommand{\thefigure}{S.\arabic{figure}}
\setcounter{section}{0}
\renewcommand{\thesection}{S.\arabic{section}}
\setcounter{table}{0}
\renewcommand{\thetable}{S.\arabic{table}}


\input{./text/SUPP.tex}

\clearpage

\let\oldthebibliography=\thebibliography
\let\oldendthebibliography=\endthebibliography
\renewenvironment{thebibliography}[1]{%
     \oldthebibliography{#1}%
     \setcounter{enumiv}{ 64 }%
}{\oldendthebibliography}

\renewcommand\refname{Supp. References}

{\small


\input{egpaper_supp.bbl}
}

\end{bibunit}

%% file: text/SUPP.tex
In this document, we present a more detailed description for the implementation of stochastic softmax sampling. We also provide experiments with decoupled sampling and pooling temperatures and additional visualizations of distributions obtained during training with softmax sampling and REINFORCE. Finally we discuss complementary experiments and results.


\section{Stochastic softmax -- Implementation}

\paragraph{Clip Sampling.}

The sampler $S$, extracts a clip of $F$ contiguous frames at temporal position $t$ from a video $x$ of arbitrary length $L$. The sampling mechanism can be formulated as: 

$S: R^{L \times 3 \times H \times W} \mapsto R^{F \times3 \times H \times W}$, with $F \leqslant L$, $H \times W$ is the spatial resolution of the data and we consider three colour channels. 
At every epoch of training, we construct batches of training clips. One clip is sampled from each training video. There are $N=L-F+1$ possible clips to extract from a given dataset example. Videos are padded to contain at least $F$ frames.
With weighted training, the temporal sampling probability distribution of each video is computed from its classification scores. In the context of a deep-learning classifier, we consider that inference class scores represent a good measure of a clip’s informativeness and relevance to the task \cite{SUPP_DBLP:conf/icmi/VielzeufPJ17,SUPP_DBLP:conf/cvpr/ZhuHSCQ16}. Specifically, for a given training clip, we use the score corresponding to the target label. In this sense, our method is similar to using the Oracle Sampler conceptualized in \cite{SUPP_DBLP:conf/iccv/KorbarTT19} at training time. This strategy minimizes the training loss by selecting the best scoring clips. The aim is to also improve the validation accuracy and reduce training time by learning from informative clips, without irrelevant and noisy frames.
Let $w_x$ be the temporal sequence of $N$ classification scores estimates corresponding to the temporal responses of the classifier convolved over $x$. Then, $w_{x,t}$  is the classification score for the training clip $x_t$. This score will be the base of our clip weighting. 

Temporal softmax sampling follows the formula:
\begin{equation}
\label{eq:weighted_sampling_SUPP}
p(S(x)=x_t) =  \frac{\exp(\gamma w_{x,t})}{\sum_{n=1}^N \exp(\gamma w_{x,n})} \; \forall x_t \subset x. 
\end{equation}


\paragraph{Distribution updates.}

At every epoch, a single clip is selected from each video in the training dataset. We apply inference on the sampled clip without data-augmentation (as this would introduce noise in the score distribution), and separately use a copy with data-augmentation to train the model. The importance of evaluating scores from ``clean" clips is further discussed in Section~\ref{sec:Supp_results}.

We employ a propagation mechanism to update several clip probabilities from a single clip evaluation. After training with clip $x_t$, obtaining classification score $f(x_t)$, we update the estimate of score $w_{x,t+i}$ for all $i$ in $[-F;F]$ with linear interpolation: 
\begin{equation}
w_{x,t+i} = w_{x,t+i} + \frac{F - \lvert i \lvert}{F}  (f(x_t) - w_{x,t+i}). 
\end{equation}


\paragraph{Training phases.}
Efficient training-clip sampling is highly dependant on the accuracy of the temporal distributions. Since the estimate $w$ is built iteratively and the model is trained simultaneously, it could take a long time for clip sampling to become interesting. Waiting for all clips to be evaluated before starting to sample them efficiently would require an unreasonable number of epochs. Also, we do not want to update distributions with classification scores obtained from an untrained model. Therefore, we implemented several mechanisms to bootstrap the distributions, to make them representative of the informativeness of clips as early as possible during training, without introducing heavy computational overhead. We decompose the training process into three simple steps relative to the sampling mechanism: first, warm-up with uniform sampling and no distribution updates, then, exploration with deterministic sampling and initialization of distributions, and finally exploitation with softmax sampling and distribution updates as described above. The number of epochs associated to each of these phases can be adapted to the task, based on the total duration of training, the average video length and the clip size.

During the first 3 epochs, uniform sampling is used without  updating the distributions. This warm-up time allows the model to jump from 14\% to 25\% validation accuracy on AFEW, which is about half of the final accuracy. 
After the 3rd epoch, the classification scores are far more reliable to update sampling distributions.

Before starting to exploit the sampling mechanism, we need to build entire temporal distributions of the videos in order to have information on the relative importance of each clip. 
One solution is to compute classification scores on the entire training videos with an inference step. This would be very expensive as a single training video can have hundreds of frames, and the number of possible $F$-frame overlapping clips to evaluate is $L-F+1$. Also, the model's temporal receptive field and the duration of training clips are not equal, so using score vectors computed convolutionally from long videos might not be reliable and wouldn't be coherent with the scores obtained from short clips in the exploitation step. Instead, we propose a lighter exploration step that integrates smoothly into our framework. 
For 5 epochs, we sample training clips deterministically from uniformly spaced temporal locations. The selected clips are used for training and provide classification scores to initialize the corresponding $\frac{1}{5}$ of the distributions. In our implementation, the score is obtained before the back-propagation. Evaluating the clip right after training on it would bias the distributions toward rewarding most fitted clips, while we aim at evaluating their informativeness. 
Overall, this exploration step enforces a diversity of clips in the early stages, which is important for building representative distributions. 

Then, for the main part of the training, training clips are sampled stochastically, based on the softmax probabilities computed from $w$. We keep updating the distributions throughout training. As clips share a lot of frames with their neighbours, we update the distribution around the sampled temporal location. We consider this particularly important in our experiments because the number of clips in videos is generally greater than the number of training epochs. We use linear interpolation centered on the selected clip and propagate to 16 frames on each side, with decreasing update weight further from the center. The settings for the number of steps in each step and the method for smoothing distributions can be optimized and adapted to the task at hand, and require more attention in future work.


\section{Discussion on sampling temperatures}

To evaluate the benefits of temporal softmax weighting, we study the impact of the joint sampling and pooling temperature parameter on the classification performance and training time. Results on AFEW are reported in Table~\ref{tab:AfewResults}. The baseline is the uniform training, with average pooling during inference. This is equivalent to the temperature $\gamma = 0$ in the softmax framework. However, our method comprises warm-up and exploration as described in the implementation discussion, so we report both the results with uniform sampling and with $\gamma = 0$. We can see that deterministic exploration seems to be beneficial as it provides more variations in training data, but consequently delays convergence by acting as a regularizer. Weighted sampling is effective after the $8^{th}$ epoch (3 epochs of warm-up and 5 epochs of exploration). With large $\gamma$ temperature parameters, the best clips are exploited more often, leading to lower training data variation and more informative clips. We observe a shortening in the duration (epochs) of the training process. The best classification performance (47.35\%) is obtained with softmax temperature $\gamma = 1$, which provides a compromise between uniform and maximum temperatures, effectively focusing on relevant clips while maintaining diversity in selection. Results with softmax temperatures during training and average pooling during testing demonstrate the effect of sampling temperature on the learning phase. Interestingly, we observe that having different temperature during training (clip sampling) and during testing (temporal pooling) can lead to even better performance. When considering this possibility, the best accuracy (47.55\%) is obtained with $\gamma_s = 1$ for training-clip sampling, and $\gamma_p = 10^6$ for softmax pooling. 
The \textit{Sampling only} experiment on UNBC-McMaster reported in Table~\ref{tab:UNBC_results_SUPP} also supports this hypothesis. It shows that weighted sampling improves training quality on its own (more details below). Although here, average pooling at test time seems more efficient than max pooling, probably due to differences between the categorical emotion recognition task and the binary pain detection setup (the benefits of temporal softmax being limited for No Pain videos). As we designed this method as a unifying framework, we do not extensively study the effect of decoupling the two softmax temperatures.

\begin{table*} 
  \begin{center} 
    \begin{tabular}{l||c|c|c|c} 
      \hline  
      \textbf{Training $\gamma_s$} & \textbf{Accuracy (\%)}  & \textbf{Epochs} & \textbf{Test $\gamma_p = 0$} & \textbf{Test $\gamma_p = 10^6$}\\

      \hline   \hline


    uniform & 45.66 $\pm 0.21$ & 24.55 $\pm 2.75$ & 45.66 & 46.91 \\
    $\gamma_s = 0$  & 46.07 $\pm 0.20$ & 25.66 $\pm 3.14$ & 46.07 & 46.86 \\
    $\gamma_s = 0.5$ & 46.07 $\pm 0.27$ & 23.56 $\pm 3.09$ & 45.61 & 47.00 \\
    $\gamma_s = 1$ & 47.35 $\pm 0.27$ & 20.33 $\pm 1.72$ & 46.59 & \textbf{47.55} \\
    $\gamma_s = 10$ & 46.65 $\pm 0.40$ & 17.22 $\pm 2.20$ & 45.84 & 46.76 \\
    
      \hline
    \end{tabular}
  \end{center}
      \caption{Results obtained by decoupling training (sampling $\gamma_s$) and testing (pooling $\gamma_p$) softmax temperatures on the AFEW dataset. Models are trained with the clip-sampling strategy indicated in the left column, and results are provided for tests with average ($\gamma_p = 0$) and max ($\gamma_p = 10^6$) video-level temporal pooling. The \textit{uniform} entry and $\gamma_s = 0$ differ because of the deterministic exploration at the beginning of training. }
    \label{tab:AfewResults}
\end{table*}

\begin{table} [h]
  \begin{center} 
    \begin{tabular}{|l|c|} 
      \hline
      \textbf{Training method} & \textbf{Acc. (\%)}\\
    \hline
    Uniform training $\gamma = 0$ & 45.87 \\
    Clean scoring, $\gamma = 10$	& 46.91  \\
    Data-augmented scoring, $\gamma = 10$ & 46.65 \\

      \hline
    \end{tabular}
  \end{center}
      \caption{Influence of scoring from clean samples compared to directly using the data-augmented training samples, on AFEW.}
    \label{tab:clean_scores}
\end{table}

\section{Additional Results and Discussions}
\label{sec:Supp_results}

\paragraph{Clean sample scoring.}
In order to update the sampling distributions, it is straightforward to use the scores obtained during training. However, these scores are subject to data-augmentation and dropout. This introduces noise in the estimation of temporal distributions. Table \ref{tab:clean_scores} shows this phenomenon.
Using ``clean" copies of the clips to evaluate their score makes temporal sampling more efficient. In our experiments, using softmax temperature $\gamma = 10$, accuracy was 46.65\% with data-augmented samples and 46.91\% with clean samples (with a uniform sampling baseline of 45.87\%). The computational overhead is very limited as it consists in adding only an inference step on small clips with no back-propagation. We can note that even when taking the readily available scores to update the distributions, our sampling method performs better than uniform sampling.
\begin{table} 
  \begin{center} 
    \begin{tabular}{|c|c|c|c|} 
    \hline
       & \multicolumn{3}{c|}{\textbf{Clip duration (frames)}} \\
      
      \textbf{$\gamma$ Temp.  } & 
      \textbf{8} & \textbf{16} & \textbf{32}\\
    \hline
    0	& 45.17 & 45.78 & 47.09 \\
    1	& 46.39 & 47.00 & 47.43 \\
    10	& 46.04  & 45.85 & 47.17 \\

      \hline
    \end{tabular}
  \end{center}
      \caption{Influence of training-clip duration for classification accuracy (\%) with 3D-CNN stochastic softmax on AFEW.}
    \label{tab:clip_duration}
\end{table}

\paragraph{Clip duration.}
We study the effect of training-clip duration on classification accuracy, for different temperatures of sampling. We perform very limited hyper-parameter search for this study, so performances could probably be improved for large clip duration. Results presented in Table~\ref{tab:clip_duration} show that accuracy improves with clip size, but the impact of stochastic softmax is greater for smaller clips. Uniform sampling ($\gamma = 0$) particularly benefits from larger clips, as they will reduce noise in gradients and training inputs will be closer to those in inference mode (long videos). Weighted sampling on the contrary has more impact with small clips, as they allow for more precise focus and avoiding of irrelevant clips. Note that all clips become similar when their size is large, with more overlapping frames.

\begin{table} 
  \begin{center} 
    \begin{tabular} {|l|c|c|} 
      \hline 

       \multirow{2}{*}{\textbf{Method}}
      & \textbf{EER} 
      & \multirow{2}{*}{\textbf{Epoch}}\\
        & \textbf{Acc. (\%)}    &\\
    \hline  \hline

        Our baseline 3D VGG (unif.) & 86.58& 43.0 \\
        Stochastic Softmax 
        ($\gamma = 2$) & 87.21 &  37.4 \\
        Sampling only ($\gamma_s = 2$) & 87.63 & 35.2  \\
        PSPI sampling ($\gamma_s = 0.8$) & 87.84  & 25.8 \\
        \hline
 
    \end{tabular}

  \end{center}
      \caption{Additional results of a 3D~CNN on UNBC-McMaster, we compare the baseline (uniform training and average pooling) with temporal stochastic softmax as proposed in the paper ($\gamma_s~=~\gamma_p~=~2$), a decoupled version of temporal stochastic softmax ($\gamma_s = 2$ and $\gamma_p = 0$) and an experiment involving expert frame-level labels to guide sampling. 
    }
    \label{tab:UNBC_results_SUPP}
\end{table}

\paragraph{Sampling with frame-level labels.}
As the UNBC-McMaster dataset provides expert-annotated PSPI scores, measuring pain intensity at each frame, it can constitute an alternative to our estimated sampling distributions. Table~\ref{tab:UNBC_results_SUPP} reports the performance of a model trained with short-clips sampled with the PSPI distributions. As the PSPI range is 0-16, much higher than the classification scores produced by the model, we use a temperature $\gamma_s = 0.8$. With an accuracy of 87.84\%, this model performs better than the weakly-supervised model. The improvement is quite limited, confirming that stochastic softmax is able to estimate meaningful distributions from sequence-level labels only.

\begin{figure}  
\centerline{\includegraphics[height=0.8\textheight]{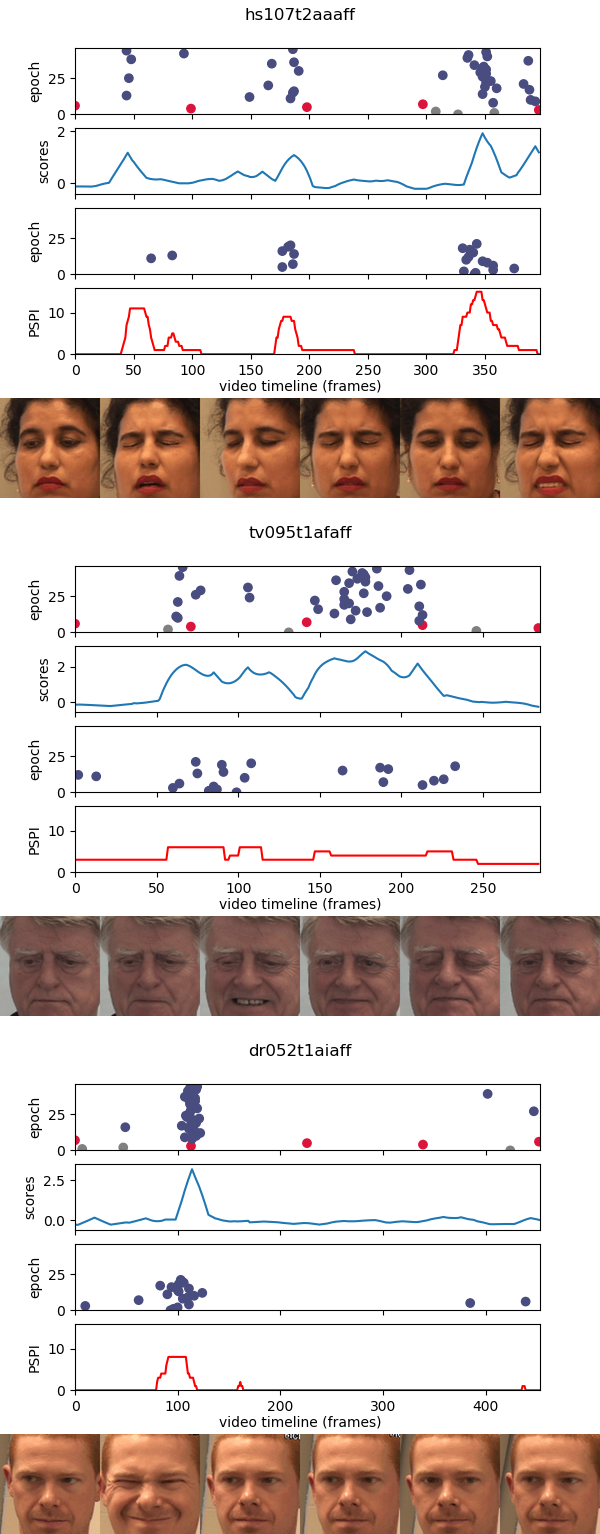}}
\caption{Visualizations of stochastic softmax training for three Pain samples of the UNBC-McMaster dataset. For each sample, the figures describe, from top to bottom, the sampling maps (temporal location for each epoch) and corresponding temporal sampling distributions, for stochastic softmax training from sequence-level labels (OPI) versus frame-level annotations (PSPI). The sequence-level label of each video is OPI~3 for \textit{hs107t2aaaff} and \textit{tv095t1afaff}, OPI~5 for \textit{dr052t1aiaff}.
}
\label{fig:unbc_print}
\end{figure}

Figure~\ref{fig:unbc_print} provides more details to compare PSPI and weakly-supervised sampling. The distributions estimated from sequence-level classification scores have clear similarities with the PSPI curves. We see that a data-oriented sampling strategy can replace the need for more labels. Theoretically the proposed method could also learn distributions for No Pain, while PSPI scores are generally zero for this class, but this doesnt seem to be relevant in our experiments.

Also, Werner \etal \cite{SUPP_DBLP:journals/taffco/WernerALWGT17} discussed limitations of the Prkaching and Solomon Pain Intensity (PSPI) scores, how they do not always correspond to pain expressions, and how their high temporal resolution might be misleading. This suggests that using expert annotations are not necessarily the best approach, even when they are available. However, a clear advantage of the PSPI-based sampling is the possibility to train with high intensity sampling directly, without exploration. In our experiments, this translates into a reduction of training time from 35.2 to 25.8 epochs in average.

\pagebreak

\section{Additional visualizations of training}

\begin{figure}
\centering
\includegraphics[width=0.8\linewidth]{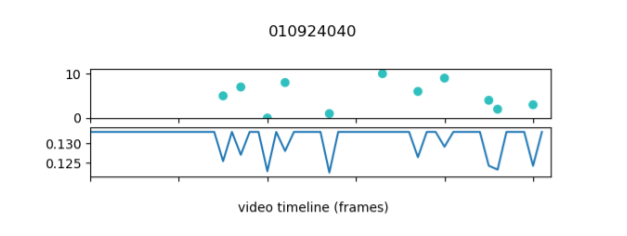}
\caption{Distribution updates obtained with REINFORCE are small and localized. The distributions take too much time to fit for the needs of training.}
\label{fig:reinf_update}
\end{figure}

Figure~\ref{fig:reinf_update} displays an example of training distribution obtain with the noisy updates of REINFORCE.

\begin{figure} 
\centerline{\includegraphics[width=0.9\linewidth]{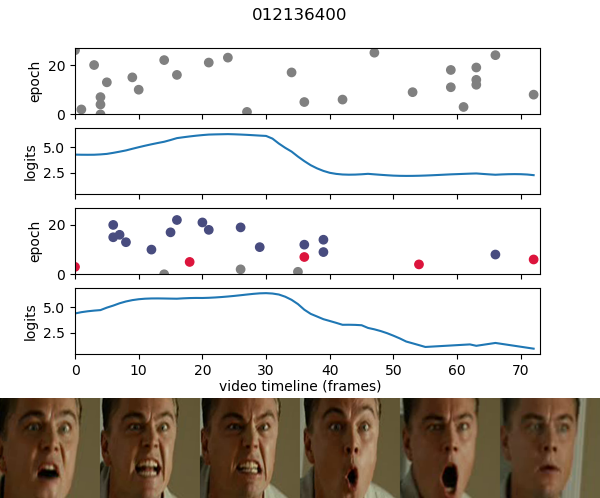}}
\caption{Visualization of sampling distributions for uniform training (above) and softmax temperature 1 (bellow), for sample 012136400 of the Angry category.}
\label{fig:Angry_012136400}
\end{figure}

\begin{figure} 
\centerline{\includegraphics[width=0.9\linewidth]{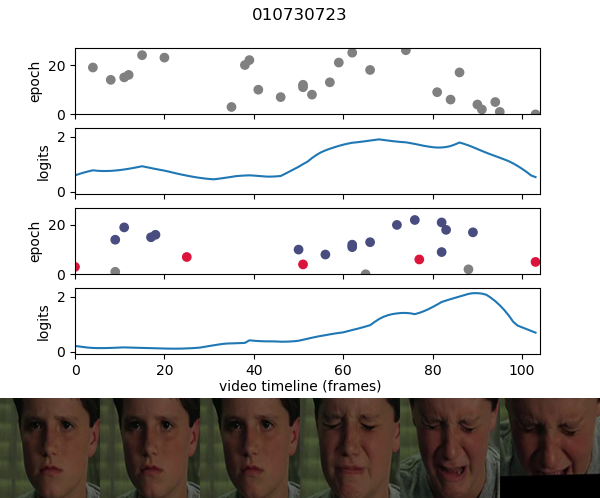}}
\caption{Visualization of training for sample 010730723 of the Sad category, with a clear emotional progression from Neutral to Sad, and occlusion in the final frames.}
\label{fig:Sad_010730723}
\end{figure}

\begin{figure} 
\centerline{\includegraphics[width=0.9\linewidth]{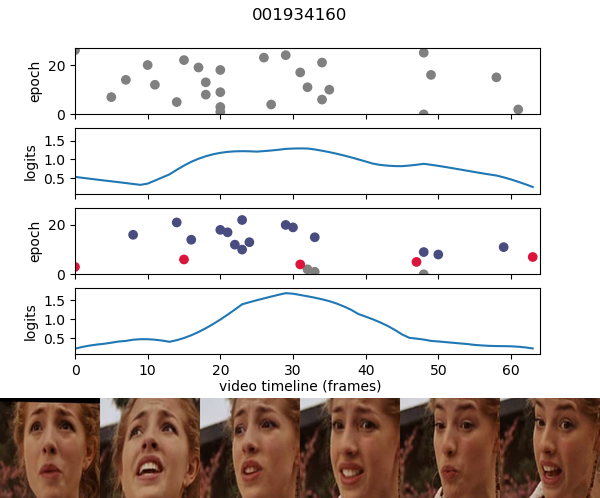}}
\caption{Visualization of training for sample 001934160 of the Disgust category, focusing on the least ambiguous expressions.}
\label{fig:Disgust_001934160}
\end{figure}

\begin{figure} 
\centerline{\includegraphics[width=0.9\linewidth]{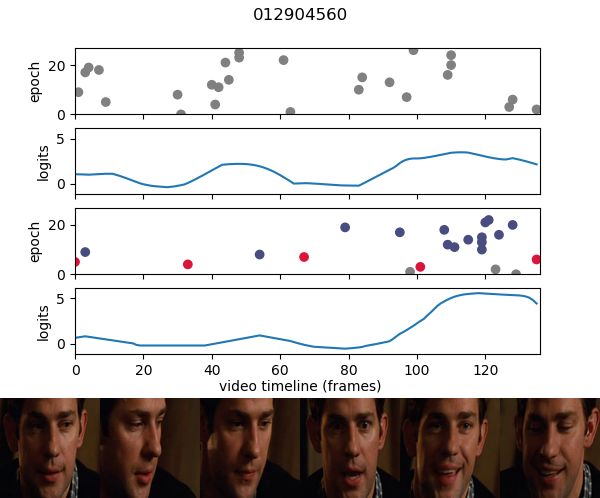}}
\caption{Visualization of training for sample 012904560 of the Happy category, avoiding Neutral and Surprise expressions to focus on the Happy frames at the end of the video.}
\label{fig:Happy_012904560}
\end{figure}

\begin{figure} 
\centerline{\includegraphics[width=0.9\linewidth]{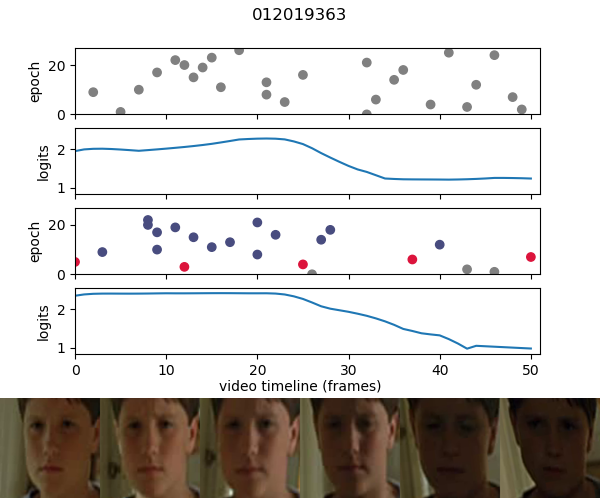}}
\caption{Visualization of training for sample 012019363 of the Neutral category, with lightning variations rendering the end of the video uninformative.}
\label{fig:Neutral_012019363}
\end{figure}

\begin{figure} 
\centerline{\includegraphics[width=0.9\linewidth]{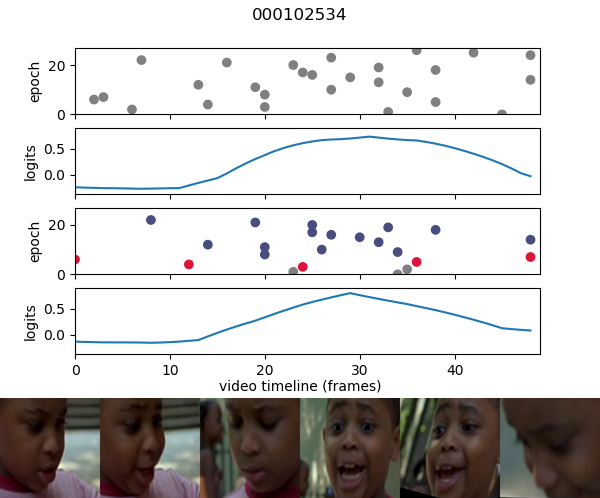}}
\caption{Visualization of training for sample 000102534 of the Surprise category, with clear apex, and head-pose variations.}
\label{fig:Surprise_000102534}
\end{figure}

\begin{figure}  
\centerline{\includegraphics[width=1\linewidth]{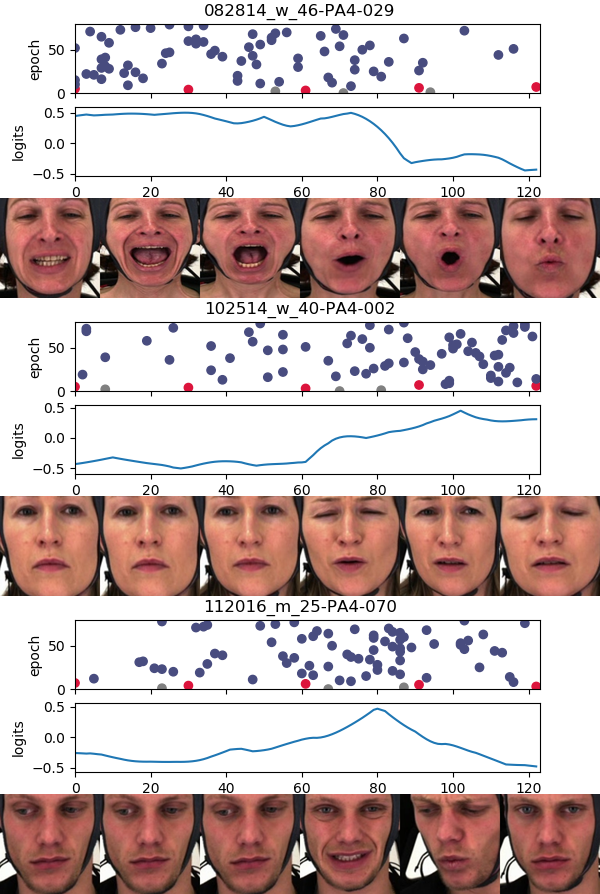}}
\caption{Visualizations of stochastic softmax training for three Pain (PA4) samples of the BioVid dataset. The inverse temperature parameter is set to $\gamma=2$.}
\label{fig:PA4_print}
\end{figure}

\begin{figure} 
\centerline{\includegraphics[width=1\linewidth]{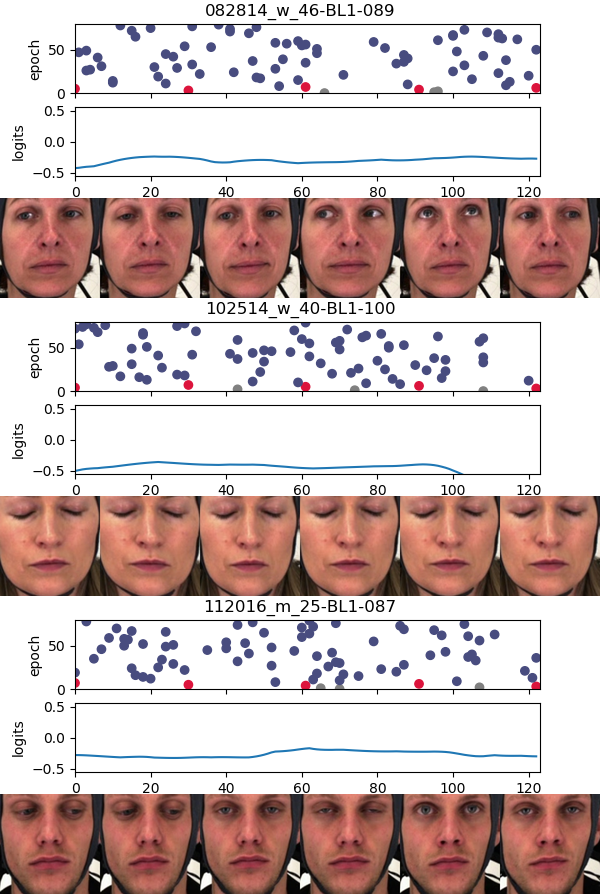}}
\caption{Visualizations of stochastic softmax training for three No Pain (BL1) samples of the BioVid dataset. The logits are small and negative, resulting in almost uniform sampling, even with high values of $\gamma$.}
\label{fig:BL1_print}
\end{figure}

Figures~\ref{fig:Angry_012136400}~to~\ref{fig:BL1_print} provides illustrative examples, for AFEW and BioVid datasets, of the temporal stochastic softmax training process. They report sampling distributions and the temporal positions of the sampled clips at each epoch of training for a specific training video. They show that the model is able to estimate meaningful distributions that correspond to the observable emotion or pain level. These distributions are built from evaluating short training clips online, iteratively through the training process.

We can note the general difference in the prediction score (logits) intensities between the datasets. This is probably due to the weight initialization of the model, which involves pretraining on 2D emotion recognition. The temperature parameter can be adapted to task-specific distributions of logits to obtain the desired sampling strategy.

On BioVid, Figure~\ref{fig:PA4_print} shows that the model is able to learn different expressions of pain.
We also visualize distributions obtained for No Pain samples of BioVid (BL1) in Figure~\ref{fig:BL1_print}. It is clear that the logits are too small to provide any real advantage over uniform sampling and average pooling for these neutral states.

\clearpage

\clearpage

\clearpage

\clearpage

\begin{table*}
  \begin{center} 
    \begin{tabular} {l|l|c} 
      \hline\noalign{\smallskip}
      \textbf{Reference} & \textbf{Architecture} & \textbf{Accuracy} \\
    \noalign{\smallskip}
    \hline
    \noalign{\smallskip}

        Li et al., 2019 
        \cite{SUPP_DBLP:conf/icmi/LiZZLTJLX19}
        & ResNet-18 + BLSTM & 43.34 \\
        2nd place AV EMOTIW 2019 
        & DenseNet-121 + BLSTM &  49.35 \\

        
        \noalign{\smallskip}
        \hline
        \noalign{\smallskip}
        
        Lu et al. 2018 
        \cite{SUPP_DBLP:conf/icmi/LuZLTLYZ18}
        & 3D VGG-16  & 39.36 \\
        3rd place AV EMOTIW 2018 
        & VGG BLSTM & 53.91 \\

        \noalign{\smallskip}
        \hline
        \noalign{\smallskip}
        
        Liu et al. 2018 \cite{SUPP_DBLP:conf/icmi/LiuTLW18}
        & 4CNNs + LSTM & 56.13 \\
        1st place AV EMOTIW 2018 
        & 3D landmarks + SVM & 39.95 \\

        \noalign{\smallskip}
        \hline
        \noalign{\smallskip}
        
        Fan et al. 2018
        \cite{SUPP_DBLP:conf/icmi/FanLL18}
        & VGG-Face & 45.16 \\ 
        2nd place AV EMOTIW 2018 
        & FG-Net & 47.00 \\

        \noalign{\smallskip}
        \hline
        \noalign{\smallskip}

        Vielzeuf et al., 2018 \cite{SUPP_DBLP:conf/icmi/VielzeufKPLJ18}
        & ResNet-18 av. pool. & 49.7\phantom{0} \\
        3rd place AV EMOTIW 2018 
        &  weighted av. pool & 50.2\phantom{0} \\

        \noalign{\smallskip}
        \hline
        \noalign{\smallskip}

        Vielzeuf et al., 2017 \cite{SUPP_DBLP:conf/icmi/VielzeufPJ17}
        & LSTM C3D 
        & 43.2\phantom{0} \\  
        4th place AV EMOTIW 2017 
        & Weighted C3D  & 42.1\phantom{0} \\  
        
        \noalign{\smallskip}
        \hline
        \noalign{\smallskip}

           Fan et al. 2016 \cite{SUPP_DBLP:conf/icmi/FanLLL16} & C3D & 39.69 \\

        \noalign{\smallskip}
        \hline
        \noalign{\smallskip}

        Bargal et al., 2016 
        \cite{SUPP_DBLP:conf/icmi/BargalBCZ16}
        & VGG-13 & 57.07 \\

        \noalign{\smallskip}
        \hline

    \end{tabular}

  \end{center}
      \caption{
    Results reported on the AFEW dataset from the literature. Differences in methodology, testing sets, use of extra-data and other factors make any comparison of these results hazardous. We provide this table as an overview of approaches and performances reported in the literature.
    }
    \label{tab:Literature_AFEW}
\end{table*}


